\documentclass[letterpaper, 10 pt, conference]{ieeeconf}  % Comment this line out if you need a4paper

\IEEEoverridecommandlockouts                              % This command is only needed if 
\usepackage[english]{babel}
\let\labelindent\relax

\usepackage{amsmath}
\usepackage{graphicx}
\usepackage[colorlinks=true, allcolors=blue]{hyperref}
\usepackage{amssymb}
\usepackage{amsthm}
\usepackage{booktabs}
\usepackage{float}
\usepackage{tikz}
\usetikzlibrary{shapes,arrows,positioning,calc}
\usepackage{multirow}
\usepackage{subcaption}
\usepackage{algorithm}
\usepackage{algpseudocode}
\usepackage[skins]{tcolorbox}
\usepackage{enumitem}
\usepackage{verbatim}
\usepackage[T1]{fontenc}
\usepackage{svg}
\usepackage{relsize}
\usepackage[rightcaption]{sidecap}
\usepackage[font=footnotesize]{caption}
\usepackage{mwe}

\title{\LARGE \bf
Learning-based Delay Compensation for Enhanced Control of Assistive Soft Robots
}
\author{Adrià Mompó Alepuz$^{1}$, Dimitrios Papageorgiou$^{1}$ and Silvia Tolu$^{1}$% <-this % stops a space
\thanks{*This work was supported by DTU Alliance fund}% <-this % stops a space
\thanks{$^{1}$Adrià Mompó Alepuz, Dimitrios Papageorgiou and Silvia Tolu are with the Department of Electrical and Photonics Engineering, Technical University of Denmark, Kongens Lyngby, Denmark {\tt\small \{amoal, dimpa, stolu\}@dtu.dk}%
        }
  \thanks{This work has been submitted to the IEEE for possible publication. 
           Copyright may be transferred without notice, after which this version 
           may no longer be accessible.}
}

\begin{document}

\maketitle
\thispagestyle{empty}
\pagestyle{empty}

\begin{abstract}
Soft robots are increasingly used in healthcare, especially for assistive care, due to their inherent safety and adaptability. Controlling soft robots is challenging due to their nonlinear dynamics and the presence of time delays, especially in applications like a soft robotic arm for patient care. This paper presents a learning-based approach to approximate the nonlinear state predictor (Smith Predictor), aiming to improve tracking performance in a two-module soft robot arm with a short inherent input delay. The method uses Kernel Recursive Least Squares Tracker (KRLST) for online learning of the system dynamics and a Legendre Delay Network (LDN) to compress past input history for efficient delay compensation. Experimental results demonstrate significant improvement in tracking performance compared to a baseline model-based nonlinear controller. Statistical analysis confirms the significance of the improvements. The method is computationally efficient and adaptable online, making it suitable for real-world scenarios and highlighting its potential for enabling safer and more accurate control of soft robots in assistive care applications.
\end{abstract}

\section{Introduction}

The growing elderly population and increasing number of individuals requiring assistance with activities of daily living (ADLs) pose significant challenges for healthcare systems worldwide \cite{who2015}. Traditional care approaches struggle to meet this growing demand, creating substantial burden on both families and caregivers \cite{McFall1992, Porta2014}. Soft robots, with their inherent safety due to material compliance, are emerging as a promising solution for direct physical assistance tasks \cite{Cianchetti2018}. Their ability to conform to irregular shapes and provide gentle interaction forces makes them particularly suitable for healthcare applications, where traditional rigid robots may pose safety concerns \cite{Rus2015}.

However, controlling soft robots presents unique challenges due to their complex nonlinear dynamics. Unlike rigid robots that can be modeled using discrete joint equations, soft robots exhibit continuous deformation along their structure \cite{Walker2013}, leading to theoretically infinite-dimensional systems. These control challenges are further exacerbated by inherent delays, which arise from communication latencies and actuator dynamics, and can significantly degrade performance or even destabilize otherwise stable systems \cite{Park2019, Zhong2006}. 
% remove the mentioning of delays in the first sentence, it is already referred in the third, remove 0.14

Although soft robot control remains an open challenge, delay compensation represents an opportunity for significant performance improvement regardless of the underlying control strategy. Traditional delay compensation methods like the Smith Predictor \cite{Watanabe1981} heavily rely on accurate system models, which are particularly difficult to obtain for soft robots. While data-driven methods for delay compensation have been explored, a key challenge remains: effectively learning these complex dynamics online to handle model changes and environment interactions, while handling efficient past signal representation from the inherent time delays.

To address these challenges, we propose a learning-based approach to approximate a nonlinear state predictor, which effectively provides measurement corrections that compensate for delays. The method combines efficient input history compression using Legendre Delay Networks (LDN) \cite{Voelker2019} with online nonlinear learning via Kernel Recursive Least Squares Tracker (KRSLT) \cite{KRSLT}. Our approach integrates seamlessly with an existing robust control framework based on super-twisting sliding mode control \cite{Papageorgiou2024}, enhancing its performance through adaptive delay compensation while preserving the robustness guarantees of the baseline controller.

 % make more generic, not too specific technicisms yet

The main contributions of this paper are: (1) the development and experimental validation of a learning-based method to approximate a nonlinear Smith Predictor for a two-module soft robot arm; (2) the demonstration of significant tracking performance improvement compared to a baseline robust controller, achieving up to 64\% reduction in tracking error at higher gains; and (3) insights into the practical implications for assistive applications.

The remainder of this paper is organized as follows. Section 2 provides relevant background in soft robot control and delay compensation. Section 3 describes the system and control architecture. Section 4 details the proposed learning-based approach. Section 5 presents the experimental setup, and Section 6 discusses the results. Finally, Section 7 concludes the paper and outlines directions for future work.

\section{Background and Related Work}

\subsection{Soft Robot Control}

Soft robots exhibit continuous deformation along their structure \cite{Walker2013}, leading to theoretically infinite-dimensional systems that are challenging to model accurately. Various control approaches have been proposed, from simplified analytical models to data-driven methods. Conventional approaches often rely on simplifying assumptions, such as the constant curvature model \cite{Hannan2003}, which become unreliable under loading conditions or complex motions. While more detailed analytical models have been proposed \cite{Falkenhahn2014}, they are often too computationally intensive for real-time control. Data-driven methods offer an alternative, with approaches ranging from neural networks \cite{Melingui2014, Thuruthel2017} to hybrid solutions that combine partial analytical knowledge with learning \cite{Reinhart2017, Papageorgiou2024}. However, these methods still face challenges to achieve both accuracy and computational efficiency for online adaptation. For a more comprehensive overview of modeling and control and methods for soft robot refer to \cite{Thuruthel2018survey, Wang2021survey}.

\subsection{State Prediction for Delay Compensation}
% be more clear about what is it approximated
The Smith Predictor \cite{Watanabe1981} is a classic approach, using a plant model to predict future outputs and adjust control. While nonlinear extensions exist \cite{Krstic2009, Fischer2013}, they remain model-dependent, a challenge for soft robots. 

Learning-based approaches offer a model-free alternative. Prior work on learning-based delay compensation has explored methods like Iterative Learning Control for linear systems or batch processes \cite{Hu2001, Hao2019}. Tian et al. \cite{Tian2015} used an offline-trained statistical method for delay prediction in linear systems. Others leverage the powerful representation capabilities of neural networks (NNs) \cite{Kim2022}, \cite{Alnajdi2023}), but rely on offline training.  Shi et al. \cite{Shi2018} presented an online nonlinear learning method based on radial basis function NNs, but its comprehensive compensation strategy introduces architectural and tuning complexity, and potential computational overhead. These limitations highlight the need for an approach that balances online adaptability, nonlinearity handling, and computational efficiency.

\subsection{Efficient Online Learning}

Kernel methods can provide this balance. Their implicit mapping to high-dimensional feature spaces allows for efficiently capturing nonlinearities \cite{Scholkopf2018, Bouvrie2017}. The Kernel Recursive Least Squares Tracker (KRLST) \cite{KRSLT} provides efficient online adaptation and computational tractability, making it well-suited for learning complex, delayed dynamics.

The challenge of representing signal history efficiently is crucial for delayed systems, where the computational burden increases with the delay length \cite{Shin1995}. The Legendre Delay Network \cite{Voelker2019} offers an efficient solution by compactly representing continuous-time signal histories using orthogonal polynomials. When combined with kernel-based learning methods, this approach becomes particularly attractive for delayed systems requiring online adaptation \cite{Liu2024}.
Our method builds on this synergy, using LDNs for efficient memory compression and KRLST for online nonlinear learning, specifically tailored to approximate the integral term of a nonlinear Smith Predictor for robust and accurate control of soft robots with delays.

The integration of such learning-based methods with traditional robust control frameworks represents a promising approach for enhancing the performance of soft robots while maintaining stability guarantees. This paper explores such an integration, focusing specifically on efficiently learning a nonlinear Smith Predictor to improve delay compensation in a two-module soft robot arm.
% remove integral thing

\begin{figure}[htbp]
    \centering
    \setlength{\fboxrule}{0pt}  % Makes the frame invisible
    \setlength{\fboxsep}{0pt}   % Removes padding inside the box
    \begin{minipage}[c]{0.55\columnwidth}
        \framebox{\parbox{0.9\columnwidth}{
            \includegraphics[width=\linewidth]{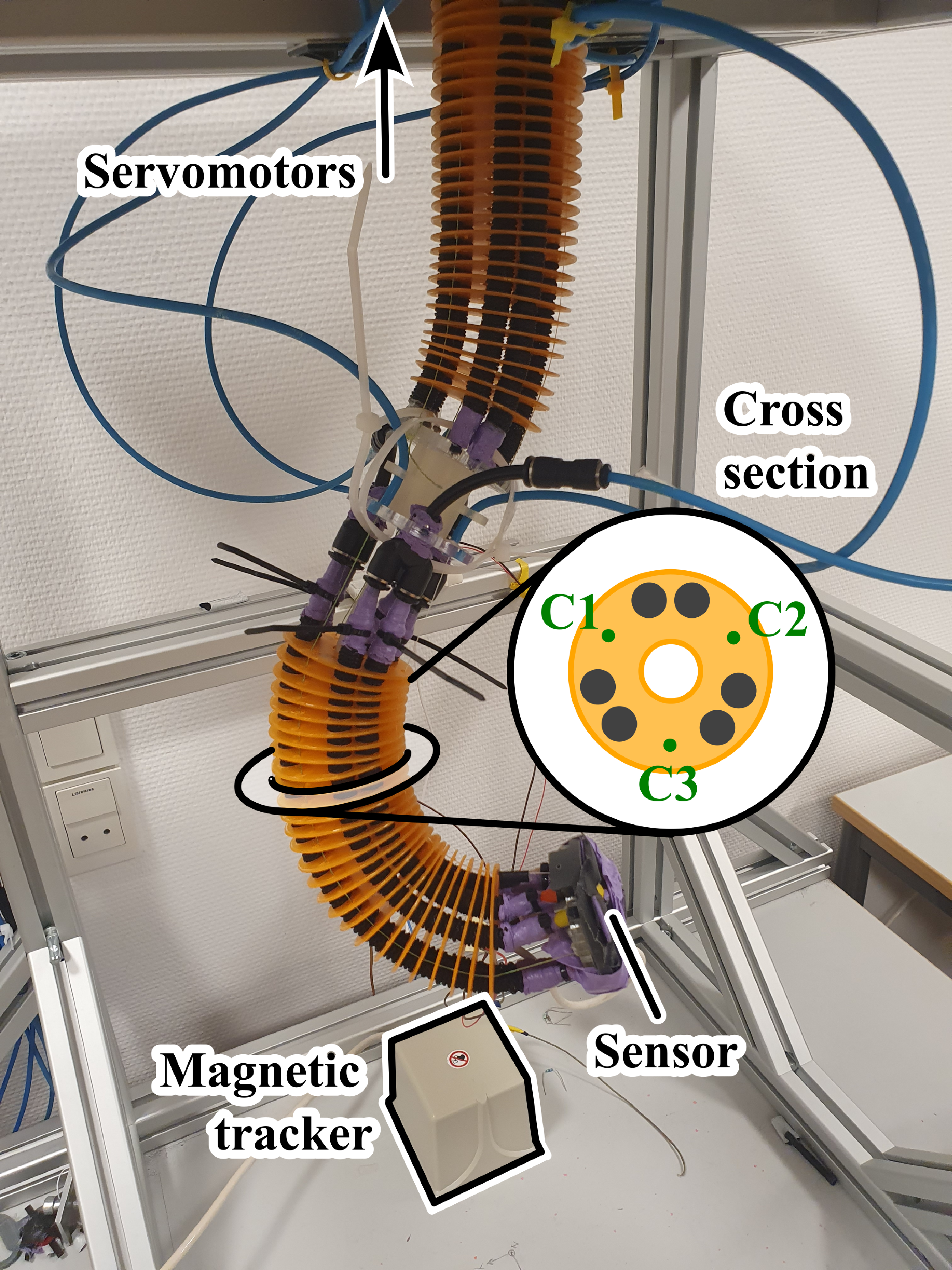}
        }}
    \end{minipage}%
    \begin{minipage}[c]{0.4\columnwidth}
        \vspace{0em}
        \caption{Soft robot arm with sensor and actuator setup. The cross-section shows the distribution of the three cables (C1, C2, C3) in one of the modules.}
        \label{fig:enter-label}
    \end{minipage}
\end{figure}

\section{System Description}

\subsection{Soft Robot Arm}

The soft robot arm consists of two identical modules connected in series, each actuated by three servomotors for cable-driven control. The use of this arm for assistive applications in elderly care was explored in \cite{ISUPPORT}. A magnetic tracker mounted at the tip provides 6-dimensional measurements (3D position and 3D orientation). This two-module configuration presents increased challenges compared to \cite{Papageorgiou2024}, with higher inertial effects and stronger dynamic coupling between modules.
% reference I-Support

\subsection{System Model} 

Following \cite{Papageorgiou2024}, we employ a first-order nonlinear model identified through SINDYc (Sparse Identification of Nonlinear Dynamics with control). While this choice omits explicit acceleration terms, it offers practical advantages: simpler identification and effective capture of dominant viscoelastic dynamics, though the increased mass in our two-module system makes this modeling choice more challenging. The identified model takes the form 
\begin{equation}
    \dot{x} = Ax + f_A(x) + [B_1 + B_2(x)]u + g(u)
    \label{eq-sys}
\end{equation}
where $x \in \mathbb{R}^6$ represents the end-effector pose and $u \in \mathbb{R}^6$ contains the servomotor inputs. The control loop operates at 50 Hz sampling rate.

\subsection{Control Architecture}

The proposed control architecture enhances the robust control framework from \cite{Papageorgiou2024} by integrating a nonlinear Smith Predictor framework to address the 0.14-second input-output delay. The baseline controller employs a super-twisting sliding mode controller (STSMC) for robustness against uncertainties and disturbances, coupled with a nonlinear input estimator to handle actuator nonlinearities.

\begin{figure}[h]
    \centering
    \setlength{\fboxrule}{0pt}  % Makes the frame invisible
    \setlength{\fboxsep}{0pt}   % Removes padding inside the box
    \framebox{\parbox{0.45\textwidth}{
        \includegraphics[width=\linewidth]{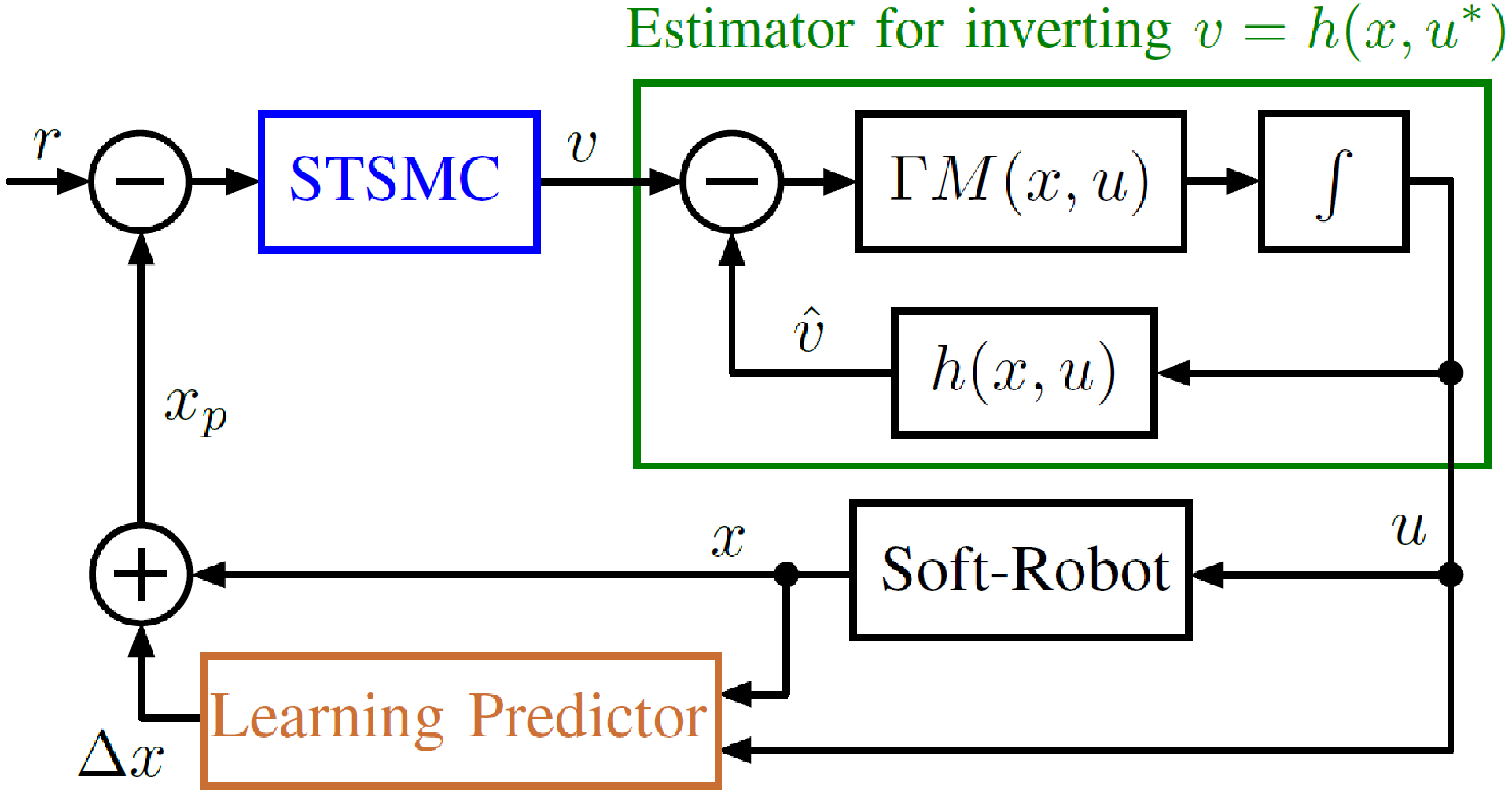}
    }}
    \caption{Control architecture combining the model-based baseline control architecture (STSMC and inversion estimator) together with the proposed Learning Predictor to realize a Smith Predictor control architecture.}
    \label{fig:ctrl_arch}
\end{figure}

The SINDYc-identified model described in Equation \ref{eq-sys} defines the nonlinear mappings needed by the controller, with $h(x,u) = [B_1 + B_2(x)]u + g(u)$ as the nonlinear input mapping, and $f(x)=Ax + f_A(x)$ as the state-dependent dynamics. The STSMC law is then given by:
\begin{equation}
% v_{SMC} = k_1 \Vert e \Vert^{1/2} \frac{e}{\Vert e \Vert + \varepsilon} + k_2 \int_0^t \frac{e(\tau)}{\Vert e(\tau) \Vert + \varepsilon} d\tau
v_{SMC} = k_1 \lfloor e \rceil^{1/2} + k_2 \int_0^t \lfloor e(\tau) \rceil^0 d\tau
\label{eq:sts_control_law}
\end{equation}
where $k_1$ and $k_2$ are positive definite matrices, $e = x - r$ is the pose tracking error and the notation $\lfloor w \rceil^a$ refers to a vector $w$ with components $w_i = \vert w_i \vert^a\text{sgn}(w_i)$. The desired control speed is then computed as $v = \dot{r}-f(x)-v_{SMC}$, and designed such that, if $h(x,u^*)$ realizes $v$ with an ideal input $u^*$, the closed-loop error dynamics become finite-time stable.

The nonlinear input estimator inverts the mapping $h(x, u)$ between the desired control speed $v$ and the actual actuator input $u$, following:
\begin{equation}
\dot{u} = \Gamma M(x,u)(v - h(x,u))
\end{equation}
where $\Gamma$ is a positive definite matrix, and $M(x, u)$ is a matrix function designed to ensure invertibility conditions.

As shown, both the commanded velocity $v$ and the input estimator depend on the velocity model. While effective for single-module control \cite{Papageorgiou2024}, this model faces limitations with the increased inertia and coupling in our two-module system, making accurate state prediction more challenging and motivating our learning-based Smith Predictor enhancement. This requires considering the full state (pose and its time derivative) despite using a velocity-based model for control.

\subsection{Nonlinear Smith Predictor}

For a delayed nonlinear system, we consider the \emph{complete state} $X(t) = [x(t), \dot{x}(t)]^T$, including pose $x(t)$ and velocity $\dot{x}(t)$. The complete system dynamics are given by $\dot{X}(t) = \mathcal{F}(X(t), u(t-d))$, with $\mathcal{F}$ as a generic nonlinear function. The SP aims to predict the future pose $x(t+d)$. Ideally, the predicted pose $x_p(t)$ is given by considering the predicted \emph{complete state} $X_p(t) = [x_p(t), \dot{x}_p(t)]^T$ within the integral:
\begin{equation}
x_p(t) = x(t) + \underbrace{\int_{t-d}^t  \mathcal{F}_x(X_p(\tau), u(\tau)) d \tau}_{\Delta x_p(t)}
\label{eq:smith_predictor_integral_approx}
\end{equation}
where $\Delta x_p(t)$ is the predicted pose change, and $\mathcal{F}_x$ represents the pose component dynamics derived from $\mathcal{F}$.  However, computing this integral iteratively using predicted states $X_p(\tau)$ for $\tau \in [t-d, t]$ is complex and error-prone, and more so with the absence of a good model for $\mathcal{F}_x$.

Instead, we aim to approximate the effect of the entire integral term. For this, we identify that the key variables it depends on are just the history of control actions $u_{[t-d, t]}$ and the initial predicted state $X_p(t-d)$. The rest of the predicted states up to $X_p(t)$ are unknown but not necessary, since they are produced by the model $\mathcal{F}$. We also establish $X_p(t-d) \approx X(t)$, leveraging the fact that an ideal predictor at $t-d$ would have predicted the state at time $t$. The signal $X(t)$ is known, as a composition of the current pose $x(t)$, and the estimated velocity $\dot{\hat{x}}(t)$ from an observer $\dot{\hat{x}} = L(x - \hat{x})$.
The nonlinear mapping method that approximates $\hat{y}(t) \approx \Delta x_p(t)$ is detailed in the next section. 

Finally, the predicted pose $x_p(t) = x(t) + \hat{y}(t)$ is used for error calculation in the STSMC controller, aiming to compensate for the delay and enhance tracking performance.

\begin{figure}[h]
    \centering
    \setlength{\fboxrule}{0pt}  % Makes the frame invisible
    \setlength{\fboxsep}{0pt}   % Removes padding inside the box
    \framebox{\parbox{0.48\textwidth}{
        \includegraphics[width=\linewidth]{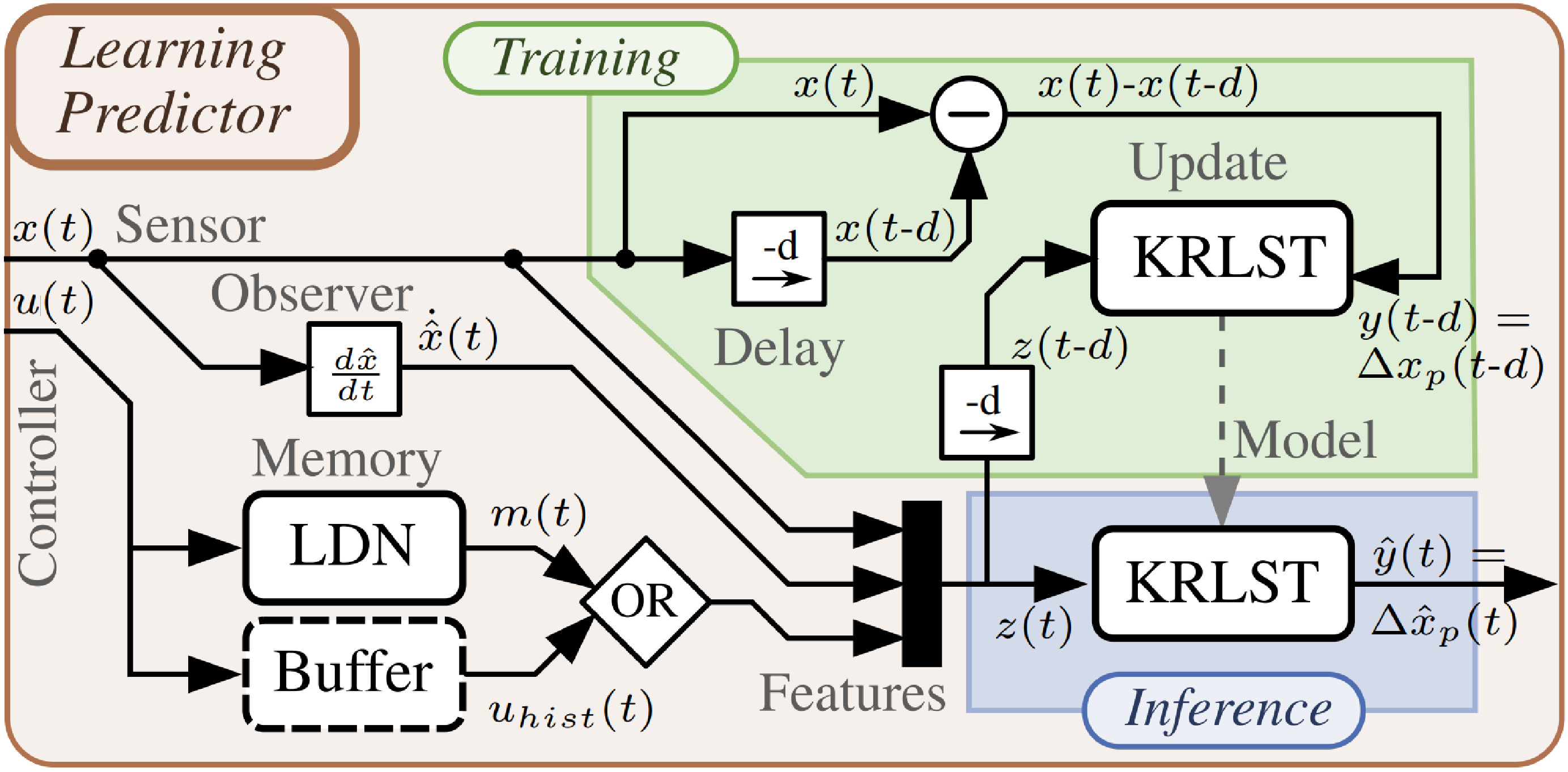}
    }}
    \caption{Learning Predictor, showing the dependency on the variables $x(t), \dot{\hat{x}}(t)$ and either $m(t)$ for LDN or $u_{hist}(t)$ for raw history, and the \emph{Training} and \emph{Inference} phases.}
    \label{fig:lp}
\end{figure}

\section{Learning-Based Approximation of the Smith Predictor Integral Term}

This section details the proposed learning-based approach for approximating the integral term of the nonlinear Smith Predictor. The method combines input history compression using Legendre Delay Networks (LDNs) with online nonlinear learning via the Kernel Recursive Least Squares Tracker (KRLST) algorithm.

\subsection{Input History Compression with LDN}

To address the challenge of representing the extended input history required for delay compensation, we employ LDNs to compress the past input signal. The LDN approximates the signal over a time window $\theta$ using a set of $p$ orthogonal Legendre polynomials, effectively capturing the essential information about the past input in a compact representation.

The dynamics of the LDN are governed by the following linear time-invariant system:
\begin{equation}
    \theta\dot{m}(t)  = \mathbf{A} \mathbf{m}(t)+\mathbf{B} u(t)
\end{equation}
where $\mathbf{m}(t) \in \mathbb{R}^p$ is the state vector, $\mathbf{A} \in \mathbb{R}^{p \times p}$ and $\mathbf{B} \in \mathbb{R}^{p \times 1}$ are matrices derived from the Padé approximation of a delay, and $u(t)$ is the input signal. The specific structure of $\mathbf{A}$ and $\mathbf{B}$ matrices can be found in \cite{Voelker2019}.

In this work, we use $p=3$ states for the LDN. This provides a balance between compression and accuracy in representing the input history. The memory length $\theta$ is set to match the delay length $d$ of the system of 0.14 seconds.

For comparison, we consider two alternative approaches that do not use LDN compression:

\begin{enumerate}
    \item Full History: The complete history of the input signal $u_{[t-d, t]}$ is used, with the length of the history window matching the delay length $d$. This amounts to the past 7 control actions for our sampling rate.
    \item Matched History: The input history is truncated to match the number of states used in the LDN compression ($p=3$), providing a direct comparison of memory compression versus raw input history with the same number of states.
\end{enumerate}

\subsection{Online Learning with KRLST}

The compressed input history $m(t)$ (or the raw input history for non-LDN variants) is combined with the current state to form the input for the KRLST algorithm \cite{KRSLT}. The complete input vector consists of the 6-dimensional pose, its 6-dimensional time derivative (obtained through the observer), and either the compressed history $m(t) \in \mathbb{R}^3$ or raw input history $u_{[t-d,t]} \in \mathbb{R}^n$ where $n$ is either 3 or 7 states per actuator. With six actuators, the resulting input dimension is $12 + 6n$, ranging from 30 dimensions for the compressed variants to 54 dimensions for the full history variant.

Given this input $z(t)$, KRLST computes the kernel vector:
\begin{equation}
k_{t} = [k(x_1, z(t)), k(x_2, z(t)), \ldots, k(x_M, z(t))]^T
\end{equation}
where $k(\cdot,\cdot)$ is the kernel function, ${x_1, x_2, \ldots, x_M}$ are the current dictionary bases, and $M$ is the dictionary size.

The predicted output $\hat{y}(t)$, which corresponds to the approximation of the integral term in the nonlinear Smith Predictor, is then computed as:
\begin{equation}
\hat{y}(t) = k_{t}^T \alpha_t
\end{equation}
where $\alpha_t$ is the weight vector.

KRLST updates its weights online and adaptively expands its dictionary to minimize prediction error and ensure efficient representation. It also incorporates a forgetting mechanism to adapt to non-stationary dynamics. Additional key parameters are the kernel width $\sigma^2$, the regularization parameter $\nu$, and the forgetting factor $\lambda$.

In this work a Gaussian kernel was used, and the parameters $\sigma^2$, $\nu$, $\lambda$ and dictionary size $M$ were selected through a systematic tuning process detailed in Section 5.

\subsection{Approximation of the Smith Predictor}

As discussed previously, analitically computing the integral $\Delta x_p(t) = \int_{t-d}^t  \mathcal{F}_x(X_p(\tau), u(\tau)) d \tau$ within the SP is impractical. The proposed method aims to learn the nonlinear mapping $\{X(t), u_{[t-d, t]}\} \rightarrow \Delta x(t)$ online using KRLST, obtaining the approximation $\hat{y}(t) \approx \Delta x_p(t)$. 

For \textbf{inference} at time $t$, the input data $\{X(t), u_{[t-d, t]}\}$ is used to compute the prediction $\hat{y}(t)$.  The LDN provides a compressed representation of the recent past input history $u_{[t-d, t]}$ for efficient inference.

For \textbf{online training} of KRLST at time $t$, we utilize \emph{past} data from time $t-d$.  The training input is $\{X(t-d), u_{[t-2d, t-d]}\}$, which are buffered values from the previous time steps. The training target is the pose difference over the delay $\Delta x_p(t-d) = x(t) - x(t-d)$.  Thus, KRLST learns to approximate the mapping 
$\hat{y}(t-d) \approx \Delta x_p(t-d)$. The complete state $X(t-d)$ includes the observer-estimated velocity. Figure \ref{fig:lp} visually represents the described method.

By learning this mapping online, KRLST efficiently approximates the Smith Predictor integral term. The learned prediction $\hat{y}(t)$ is used to compute the predicted pose $x_p(t) = x(t) + \hat{y}(t)$ for delay compensation, bypassing complex modeling and enabling online adaptation.

\section{Experimental Setup}

The experimental validation focused on trajectory tracking performance in both transient and steady-state conditions. Each experiment consisted of tracking a circular reference path in the XY plane while maintaining constant height. The trajectory parameters were selected to ensure operation well within the robot workspace, with a 5 cm radius for the circle and the height set in middle of the vertical range.

\subsection{Experimental Protocol}
Each experiment lasted 60 seconds, structured in two phases:
\begin{itemize}
    \item A 20-second spiral buildup where the circle diameter gradually increased from zero to its final value
    \item A 40-second period of constant-diameter circular tracking at 0.5 rad/s (one revolution every 12.56s)
\end{itemize}

For analysis purposes, the data was segmented into:
\begin{itemize}
    \item Transient phase: First 22.3 seconds, including the spiral buildup
    \item Stable phase: Last 37.7 seconds, comprising the last three complete circle revolutions
\end{itemize}

Pose measurements were obtained at 50 Hz through a magnetic tracking system, with the sensor mounted at the robot end-effector. The measured 6D pose (position and orientation) and its time derivative, estimated through the observer described in Section 3, served as inputs for both control and modeling performance evaluation.

\subsection{Controller Configurations and Algorithm Tuning}
The control parameters were tuned experimentally, starting with a conservative baseline configuration that prioritized stability over tracking accuracy. The input estimator gain $\Gamma$ and sliding mode gains $k_1$, $k_2$ were adjusted to achieve smooth motion with minimal oscillations, though allowing some tracking error. This established the low gain condition, from which medium and high gain conditions were derived by doubling and tripling $k_1$ respectively:

\begin{itemize}
    \item Low gain: Conservative tuning prioritizing baseline stability, and allowing under-performance in tracking
    \item Medium gain: Doubled gain for balanced performance-stability trade-off
    \item High gain: Tripled gain to evaluate performance limits
\end{itemize}

The KRLST parameters were tuned through a two-stage process. First, offline analysis with pre-recorded data established the normalization parameters for each of the KRLST inputs, and suitable orders of magnitude for the kernel width $\sigma^2$, regularization parameter $\nu$, and forgetting factor $\lambda$. These parameters were then refined through online experiments, with adjustments of 20-50\% around the initial values to optimize performance. The dictionary size was limited to 80 entries to maintain computational efficiency while ensuring adequate model complexity.

%The KRLST algorithm parameters were tuned through preliminary offline analysis. A Gaussian kernel was used with width parameter $\sigma^2$ selected based on the input data range. The normalization parameters were set to 100, 40, and 20 for the three LDN states in LDN-3, and 200 for the input histories in Hist-3 and Hist-7. These values were determined by considering the range of the signals and further refined through offline tuning.

\subsection{Performance Evaluation}

Two primary metrics were used to evaluate the methods:
\begin{itemize}
    \item Tracking Performance: RMS error in the XY plane between reference and actual trajectories
    \item Modeling Performance: RMS error between predicted and actual state changes over the delay period, measuring the accuracy of the Smith Predictor integral term approximation. 
\end{itemize}

To quantify the benefit of prediction, the No-Pred case (no prediction) is included in the modeling performance metrics, calculated as the difference between current and future states from the experimental data of the different learning methods.

For statistical analysis, 20 successful experiments were retained for each combination of method (LDN-3, Hist-3, Hist-7) and gain condition (Low, Medium, High), after excluding runs that exhibited anomalous behavior due to sporadic computer lag or external disturbances.

Statistical significance was assessed using one-way ANOVA followed by Tukey's post-hoc test ($\alpha = 0.05$), analyzed separately for each gain condition and experimental phase.

\section{Results and Discussion}

\begin{figure*}[ht]
    \centering
    \setlength{\fboxrule}{0pt}  % Makes the frame invisible
    \setlength{\fboxsep}{0pt}   % Removes padding inside the box
    \framebox{\parbox{0.9\textwidth}{
        \includegraphics[width=\linewidth]{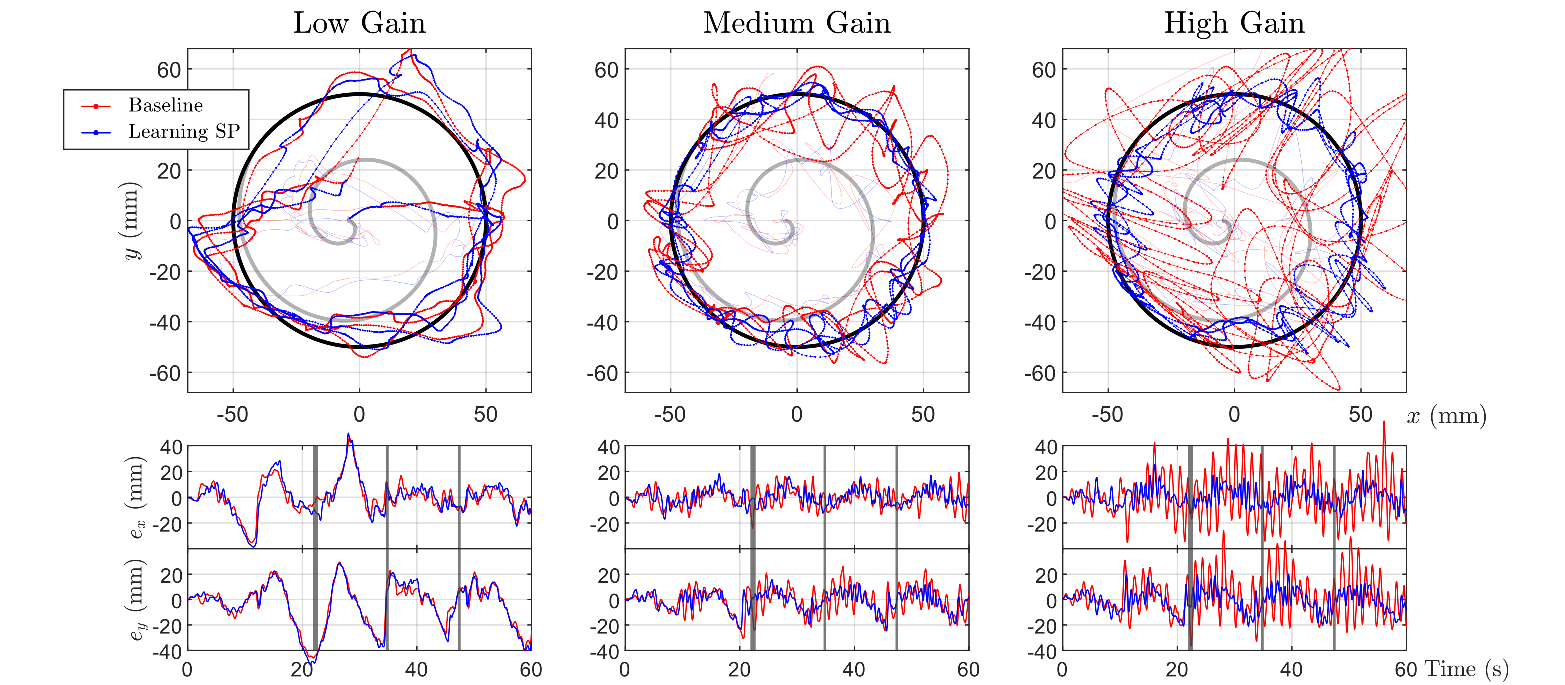}
    }}
    \caption{Comparison of XY tracking error (RMS) between baseline controller (red) and LDN-3 learning method (blue) across the gain conditions. Top plots show the trajectories in the x-y plane, with the spiral buildup faded-out; bottom plots show the x and y tracking errors that correspond to the top plots, with transient phase (first 22.3s) and stable phase (last 37.7s)}
    \label{fig:tracking_comparison}
\end{figure*}
% remove spiral, indicate top/low clarifications

\subsection{Overall Tracking Performance Comparison}

The tracking performance results demonstrate significant improvements achieved by the learning-based Smith Predictor compared to the baseline controller, particularly at higher gains. Table \ref{tab:tracking_perf} summarizes the RMS tracking error in the XY plane for both transient and stable phases.

\begin{table}[t]
\centering
\caption{XY RMS Tracking Error (mm) for Transient and Stable Phases}
\label{tab:tracking_perf}
\begin{tabular}{llccc}
\toprule
Phase & Method & Low Gain & Medium Gain & High Gain \\
\midrule
\multirow{4}{*}{Transient} 
 & Baseline & 22.92 ± 0.75 & 11.11 ± 0.46 & 15.94 ± 0.75 \\[0pt]
\cline{2-5}\\[-8pt]
 & LDN-3    & 21.78 ± 2.33 & \textbf{9.19} ± 0.68 & 9.84 ± 0.34 \\
 & Hist-3   & 22.40 ± 0.69 & 9.54 ± 0.24 & 9.97 ± 0.34 \\
 & Hist-7   & 22.17 ± 0.87 & 9.46 ± 0.27 & 10.03 ± 0.37 \\
\midrule
\multirow{4}{*}{Stable} 
 & Baseline & 21.59 ± 0.75 & 14.36 ± 0.65 & 27.63 ± 1.56 \\[0pt]
\cline{2-5}\\[-8pt]
 & LDN-3    & 20.94 ± 1.25 & 8.26 ± 0.32 & 9.88 ± 0.37 \\
 & Hist-3   & 21.37 ± 1.29 & 8.28 ± 0.30 & 9.97 ± 0.23 \\
 & Hist-7   & 21.50 ± 0.89 & 8.25 ± 0.29 & 9.92 ± 0.21 \\
\bottomrule
\end{tabular}
\end{table}
% highlight in bold

At low gains, all methods achieved similar performance, with XY RMS errors around 22 mm. However, as the controller gains increased, the advantages of the learning-based methods became more pronounced. During the stable phase, at medium gains, the learning methods achieved a 42\% reduction in tracking error compared to the baseline (8.3 mm vs 14.4 mm RMS). The relative improvement was even more dramatic at high gains, where the learning methods maintained good performance with a 64\% reduction in tracking error (9.9 mm vs 27.6 mm RMS).

During the transient phase at medium gains, LDN-3 showed a slight advantage over other variants, achieving 9.19 mm RMS error compared to 9.54 mm and 9.46 mm for Hist-3 and Hist-7 respectively. While this difference approached statistical significance (p = 0.0720), all learning methods converged to similar performance levels in the stable phase (8.25-8.28 mm RMS). This suggests that the LDN-based compression might facilitate faster initial learning, though the long-term performance is comparable across variants.

The baseline controller showed significantly different behavior between phases, performing better during the transient phase at high gains (15.94 mm vs 27.63 mm RMS), likely due to the slower motion during the spiral buildup. In contrast, the learning methods maintained consistent performance across phases, demonstrating their ability to handle both transient and steady-state tracking tasks.

Statistical analysis confirms the significance of these improvements. One-way ANOVA showed no significant differences between methods at low gains (p = 0.1787) compared to baseline. However, at medium and high gains, the differences were highly significant (p < 0.0001), with all learning variants significantly outperforming the baseline controller according to Tukey's post-hoc tests.

\subsection{Modeling Comparison of Learning Variants}

While all learning variants achieved similar tracking performance in the stable phase, their modeling accuracy and learning behavior showed notable differences. Table \ref{tab:modeling_perf} presents the modeling error results, which measure how accurately each method predicts the state change over the delay period.

\begin{table}[t]
\centering
\caption{XY RMS Modeling Error (mm) for Transient and Stable Phases}
\label{tab:modeling_perf}
\begin{tabular}{llccc}
\toprule
Phase & Method & Low Gain & Medium Gain & High Gain \\
\midrule
\multirow{3}{*}{Transient} 
 & No-Pred  & 4.26 ± 0.08 & 5.40 ± 0.14 & 7.54 ± 0.27 \\[0pt]
\cline{2-5}\\[-8pt]
 & LDN-3    & 1.45 ± 0.13 & \textbf{2.33} ± 0.15 & \textbf{2.61} ± 0.15 \\
 & Hist-3   & 1.43 ± 0.12 & 2.52 ± 0.20 & 2.85 ± 0.14 \\
 & Hist-7   & 1.41 ± 0.11 & 2.52 ± 0.14 & 2.89 ± 0.17 \\
\midrule
\multirow{3}{*}{Stable} 
 & No-Pred  & 4.24 ± 0.08 & 5.38 ± 0.13 & 7.49 ± 0.21 \\[0pt]
\cline{2-5}\\[-8pt]
 & LDN-3    & 2.21 ± 0.20 & \textbf{2.35} ± 0.10 & 3.12 ± 0.13 \\
 & Hist-3   & 2.30 ± 0.09 & 2.55 ± 0.14 & 3.28 ± 0.14 \\
 & Hist-7   & 2.38 ± 0.13 & 2.51 ± 0.11 & 3.17 ± 0.12 \\
\bottomrule
\end{tabular}
\end{table}

At low gains, all variants showed similar modeling error during the transient phase (1.41-1.45 mm RMS). However, as the gains increased, LDN-3 demonstrated superior prediction accuracy compared to both Hist-3 and Hist-7. This advantage was particularly evident at high gains during the transient phase, where LDN-3 achieved 2.61 mm RMS error compared to 2.85 mm and 2.89 mm for Hist-3 and Hist-7 respectively, a statistically significant improvement (p < 0.001).

The stable phase showed a general increase in modeling errors across all methods, likely due to the more dynamic nature of the constant-diameter circular tracking compared to the spiral buildup. However, LDN-3 maintained its advantage, particularly at medium gains where it achieved 2.35 mm RMS error compared to 2.55 mm for Hist-3 (p < 0.001).

Interestingly, increasing history length from 3 to 7 states (Hist-3 vs. Hist-7) did not significantly improve modeling or tracking. This suggests that the most recent input history (as in Hist-3) holds the most relevant information, which LDN effectively compresses. This is particularly noteworthy given that LDN-3 achieves slight superior performance while maintaining the same input dimension as Hist-3.

\begin{figure}[t]
    \centering
    \setlength{\fboxrule}{0pt}  % Makes the frame invisible
    \setlength{\fboxsep}{0pt}   % Removes padding inside the box
    \framebox{\parbox{0.49\textwidth}{
        \includegraphics[width=\linewidth]{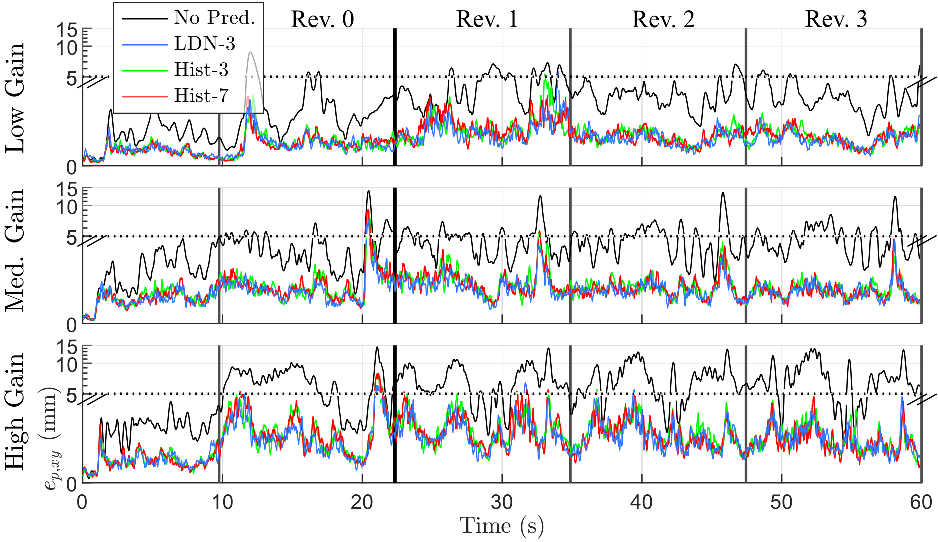}
    }}
    \caption{Comparison of XY prediction error magnitude across gain settings and methods, averaged from all 20 experiments. Error shown for LDN-3 (blue), Hist-3 (green), Hist-7 (red) methods and the No-Pred case (black). Entire revolutions are indicated as Rev. "R", with "R" = \{0,1,2,3\}. Vertical axis shows a combination of a linear scale (0-5 mm) and log scale (5-15 mm)}
    \label{fig:modeling_comparison}
\end{figure}

Figure \ref{fig:modeling_comparison} illustrates the effectiveness and adaptability of the learning methods. The initial low errors during the spiral buildup (0-20s) increase at the transition to full circular motion (Rev. 0-1). The No-Pred case consistently shows high errors (5-10mm), while learning methods achieve 2-3x lower errors. By the final revolution (Rev. 3), all learning variants converge to their lowest errors, demonstrating continuous improvement. Although stable-phase errors increase with controller gain (Table \ref{tab:modeling_perf}), this increase is less than proportional, demonstrating the robustness of the learning approach.

The relationship between modeling error and tracking performance reveals some interesting patterns. Despite the better modeling accuracy of LDN-3, all variants achieved similar tracking performance in the stable phase. The prediction accuracy achieved by all methods is sufficient for improved control performance, indicating the robustness of the control architecture to minor prediction differences. However, the improved modeling of LDN-3 during the transient phase may explain its slightly advantage at tracking, particularly at medium gains.

\subsection{Implications for Assistive Applications}

The improved tracking performance and robustness of the learning-based method have important implications for assistive applications of soft robots. In healthcare, tasks like bathing assistance \cite{ISUPPORT} require precise, reliable control and inherent safety, provided by the soft structure of the robot. The ability to achieve accurate tracking at higher gains, without large oscillations, suggests this method could enable more responsive and stable human-robot interactions.

Preliminary qualitative experiments using human-guided trajectories, where the robot follows hand movements detected by a camera, indicate smoother motion with the learning-based method compared to the baseline controller. This is relevant for assistive scenarios requiring responses to unpredictable human movements. A comprehensive evaluation of human-robot interaction is underway to validate these observations.

The efficiency of the method, especially with LDN-based compression of the input history, is particularly well suited for real-time adaptation to changing interaction dynamics. This adaptability is crucial in assistive care where the robot must adjust to different users and varying task requirements.

\section{Conclusion}

This paper has presented a learning-based approximation of the integral term in a nonlinear Smith Predictor for enhanced control of a soft robotic arm. The method combines efficient input history compression using Legendre Delay Networks with online nonlinear learning via KRLST, achieving significant improvements in tracking performance compared to a baseline robust controller.

Experimental results showed up to a 64\% reduction in tracking error at high gains, with the learning-based method effectively compensating for the system's inherent delay. The LDN-based compression variant demonstrated advantages during the learning phase, achieving superior modeling accuracy and computational efficiency. The ability of the method to enable stable operation at higher gains with reduced oscillations is relevant for assistive care, where precise and safe control are essential.

Several promising directions for future work emerge. While current results show significant improvement over the baseline, further refinements to the control model could enhance performance. The Smith Predictor effectively compensates for delays regardless of the base controller, suggesting that combining it with more sophisticated control could further reduce errors. This could include a refined, advanced model-based controller that better captures the soft robot's nonlinear dynamics, or learning-based controllers that adapt to changing conditions. Additionally, a comprehensive evaluation of human-robot interaction is ongoing to validate the effectiveness of the method in unstructured environments typical of assistive care.

The results presented here represent an important step toward enabling precise and reliable control of soft robots in healthcare settings, while maintaining their inherent safety advantages through compliant structures and learning-based adaptation.
% review acronyms in general and consistent nomenclature

\addtolength{\textheight}{-12cm}   % This command serves to balance the column lengths

%%%%%%%%%%%%%%%%%%%%%%%%%%%%%%%%%%%%%%%%%%%%%%%%%%%%%%%%%%%%%%%%%%%%%%%%%%%%%%%%

%%%%%%%%%%%%%%%%%%%%%%%%%%%%%%%%%%%%%%%%%%%%%%%%%%%%%%%%%%%%%%%%%%%%%%%%%%%%%%%%

%%%%%%%%%%%%%%%%%%%%%%%%%%%%%%%%%%%%%%%%%%%%%%%%%%%%%%%%%%%%%%%%%%%%%%%%%%%%%%%%
%\section*{APPENDIX}


\section*{Acknowledgments}

The project is gratefully funded by the DTU Alliance PhD fund.

\bibliographystyle{IEEEtran}
\bibliography{refs}

\end{document}